\def\year{2018}\relax
\documentclass[letterpaper]{article} 

\usepackage{aaai18}  
\usepackage{times}  
\usepackage{helvet}  
\usepackage{courier}  
\usepackage{url}  
\usepackage{graphicx}  

\usepackage{multirow}  
\usepackage{booktabs}  
\usepackage{amsmath}  
\usepackage{amsfonts}
\usepackage{amssymb}
\usepackage[dvipsnames,usenames]{color}

\begin{document}

\title{Co-attending Free-form Regions and Detections with Multi-modal Multiplicative Feature Embedding for Visual Question Answering}


\author{Pan Lu$^{\dagger}$~~~~Hongsheng Li$^\ddag$\thanks{Co-corresponding authors.}~~~Wei Zhang$^\natural$~~~~Jianyong Wang$^\dagger$~~~~Xiaogang Wang$^\ddag$\footnotemark[1]\\
	$^\dagger$ Tsinghua National Laboratory for Information Science and Technology (TNList)\\
	Department of Computer Science, Tsinghua University\\
	$^\ddag$ Department of Electronic Engineering, The Chinese University of Hong Kong\\
	$^\natural$ Shanghai Key Laboratory of Trustworthy Computing, East China Normal University\\
	\{lupantech, zhangwei.thu2011\}@gmail.com,~~
	\{hsli, xgwang\}@ee.cuhk.edu.hk,~~
	jianyong@mail.tsinghua.edu.cn
}
\maketitle

\begin{abstract}

Recently, the Visual Question Answering (VQA) task has gained increasing attention in artificial intelligence. Existing VQA methods mainly adopt the visual attention mechanism to associate the input question with corresponding image regions for effective question answering. The free-form region based and the detection-based visual attention mechanisms are mostly investigated, with the former ones attending free-form image regions and the latter ones attending pre-specified detection-box regions. We argue that the two attention mechanisms are able to provide complementary information and should be effectively integrated to better solve the VQA problem. In this paper, we propose a novel deep neural network for VQA that integrates both attention mechanisms. Our proposed framework effectively fuses features from free-form image regions, detection boxes, and question representations via a multi-modal multiplicative feature embedding scheme to jointly attend question-related free-form image regions and detection boxes for more accurate question answering. The proposed method is extensively evaluated on two publicly available datasets, COCO-QA and VQA, and outperforms state-of-the-art approaches. Source code is available at \url{https://github.com/lupantech/dual-mfa-vqa}.

\end{abstract}


\section{Introduction}

In recent years, multi-modal learning for language and vision has gained much attention in artificial intelligence. Great progress has been achieved for different tasks including image captioning \cite{karpathy2015deep}, visual question generation \cite{mostafazadeh2016generating,li2017visual}, video question answering \cite{ye2017video} and text-to-image retrieval \cite{xie2016online,li2017person}. The Visual Question Answering (VQA) \cite{antol2015vqa} task has recently emerged as a more challenging task. The algorithms are required to answer natural language questions about a given image's contents. Compared with the conventional visual-language tasks such as image captioning and text-to-image retrieval, the VQA task requires the algorithms to have a better understanding on both the input image and question in order to infer the answer.

\begin{figure}[!]
	\centering
	\includegraphics[width=1.0\linewidth]{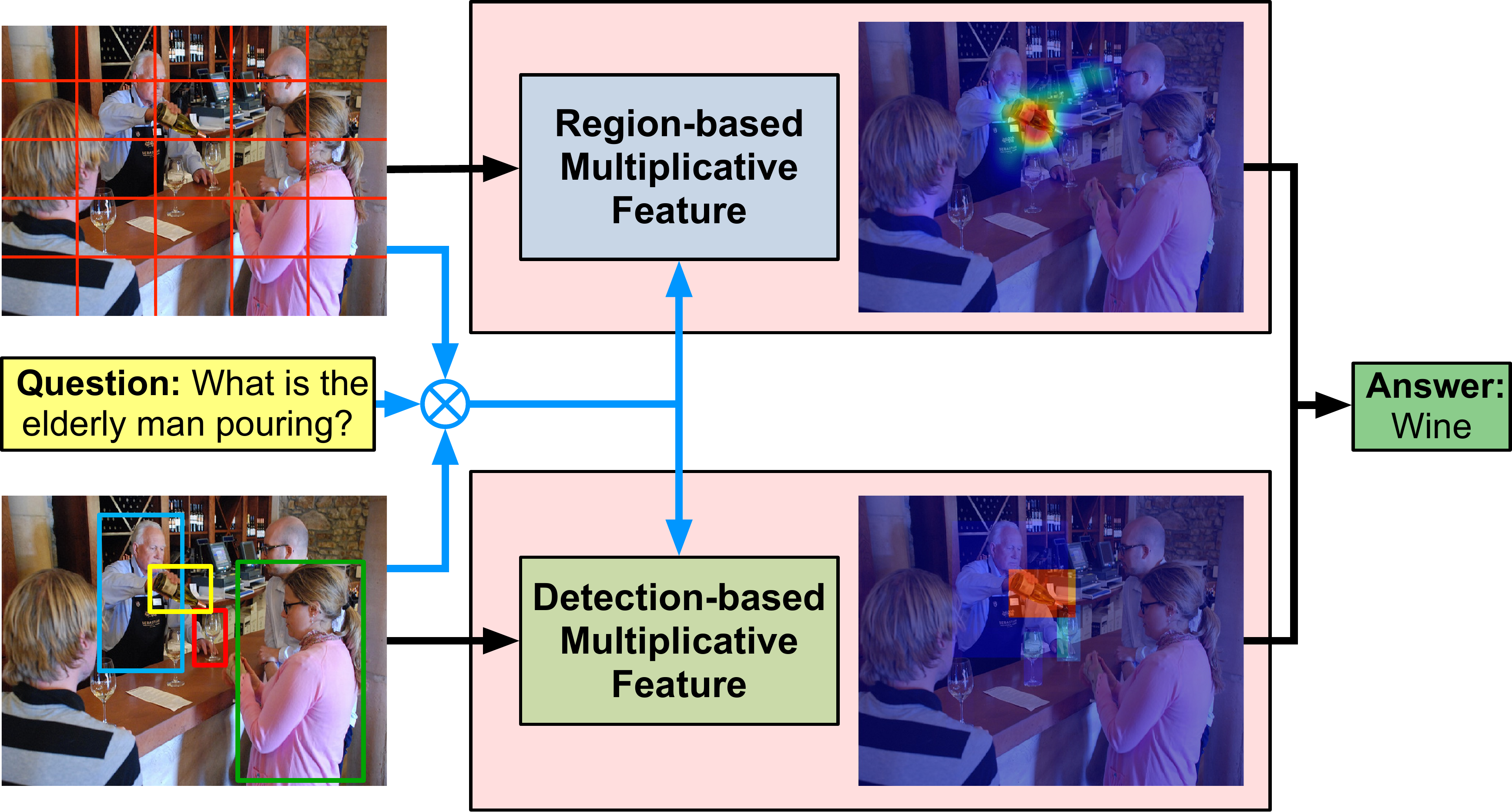}
	\caption{Co-attending free-form regions and detection boxes based on the question, the whole image, and detection boxes for better utilizing complementary information to solve the VQA task.}
	\label{fig:bimlt-demo}
\end{figure}

State-of-the-art VQA approaches utilize visual attention mechanism to relate the question to meaningful image regions for accurate question answering. Most visual attention mechanisms in VQA can be categorized into free-form region based methods \cite{lu2016hierarchical,fukui2016multimodal} and detection-based methods \cite{li2016visual,shih2016look}. For the free-form region based methods, the question features learned by a Long Short-Term Memory (LSTM) network and the image features learned by a Convolutional Neural Network (CNN) are fused by either addictive, multiplicative or concatenation operations at every image spatial location. The free-form attention map is obtained by applying a softmax non-linearity operation across the fused feature map. Since there is no restriction on the obtained attention map, the free-form attention region is able to attend both global visual context and specific foreground objects for inferring answers. However, since there is no restriction, the free-form attentive regions might focus on partial objects or irrelevant context sometimes. For instance, for an question, ``\textit{What animal do you see?}'', a free-form region attention map might mistakenly focus on only a part of the foreground cat and generate an answer of ``\textit{dog}''. On the other hand, for the detection-based attention methods, the attention mechanism is utilized for relating the question to pre-specified detection boxes (e.g., Edge boxes \cite{zitnick2014edge}). Instead of applying the softmax operation over all image spatial locations, the operation is calculated over all detection boxes. Therefore, the attended regions are restricted to pre-specified detection-box regions and such question-related regions could be more effective for answering questions about foreground objects. However, such restrictions also cause challenges for other types of questions. For instance, for the question ``\textit{How is the weather today?}'', there might not exist a detection box in the sky, resulting in a failure in answering this question.

For better understanding the question and image contents and their relations, a good VQA algorithm needs to identify global scene attributes, locate objects, identify object attributes, quantity and categories to make accurate inference. We argue that the above mentioned two types of attended mechanisms provide complementary information and should be effectively integrated in a unified framework to take advantages of both the attended free-form regions and attended detection regions. Take the above mentioned two questions as examples, the question about animal could be more effectively answered with detection-based attention maps, while the question about the weather can be better answered with the free-form region based attention maps.

In this paper, we propose a novel dual-branch deep neural network for solving the VQA problem that combines free-form region based and detection-based attention mechanisms (see Figure \ref{fig:bimlt-demo}). The overall framework consists of two attention branches, each of which associates the question with the most relevant free-form regions or with the most relevant detection regions in the input image. For obtaining more question-related attention weights for both types of regions, we propose to learn the joint feature representation of the input question, the whole image, and the detection boxes with a multiplicative feature embedding scheme. Such a multiplicative scheme does not share parameters between the two branches and is shown to result in more robust answering performance than existing methods.

The main contributions of our work can be summarized as twofold.
\begin{itemize}
	\item We propose a novel dual-branch deep neural network that effectively integrates the free-form region based and detection-based attention mechanisms in a unified framework;
	\item In order to better fuse features from different modalities, a novel multiplicative feature embedding scheme is proposed to learn joint feature representations from the question, the whole image, and the detection boxes.
\end{itemize}

\section{Related work}

State-of-the-art VQA algorithms are mainly based on deep neural networks for learning visual-question features and for predicting the final answers. Due to the establishment of the VQA dataset and the online evaluation platform by \cite{antol2015vqa}, there is an increasing number of VQA algorithms proposed every year.

\textbf{Multi-modal feature embedding for VQA}. The existing joint feature embedding methods for VQA typically combine the visual and question features learned by deep neural networks and then solve the task as a multi-class classification problem. \cite{zhou2015simple} proposed a simple baseline, which learns image features with CNN and question features from LSTM, and concatenated these two features to predict the answer.
Instead of using LSTM for learning question representations, \cite{noh2016image} used GRU \cite{cho2014learning} and \cite{ma2016learning} trained CNN for question embedding.
Different from the above mentioned methods addressing the VQA task as a classification problem, the work by \cite{malinowski2015ask} fed both image CNN features and question representations into an LSTM to generate the answer by sequence-to-sequence learning.
There are also high-order approaches for multi-modal feature embedding. MCB \cite{fukui2016multimodal} and MLB \cite{kim2016hadamard} both designed  bilinear pooling approaches to learn multi-modal feature embedding for VQA. In \cite{benyounescadene2017mutan}, a generalized multi-modal pooling framework is proposed, which shows that MCB and MLB are its special cases. Inspired by ResNet \cite{he2016deep}, \cite{kim2016multimodal} proposed element-wise multiplication for the joint residual mappings.

\textbf{Attention mechanism for VQA}. A large quantity of recent works focused on incorporating attention mechanisms for solving the VQA task.
In \cite{xu2016ask}, the attention weights over different image regions are calculated based on the semantic similarity between the question and  image regions, and the updated question features are obtained as the weighted sum of the different regions.
\cite{yang2016stacked} proposed a multi-stage attention framework, which stacks the attention modules to search question-related image regions by iterative feature fusion. The attention mechanism is not limited to image modality, \cite{lu2016hierarchical} proposed a co-attention mechanism that simultaneously attends both the question and image with joint visual-question feature representations.

Since some of the questions are related to objects in the images, object detection results are explored to replace the visual features obtained from the whole-image region. \cite{shih2016look} utilized the attention mechanism to generate visual features from 100 detection boxes for the VQA task. Similarly, \cite{li2016visual} proposed a framework that updates the joint visual-question feature embedding by iteratively attending the top detection boxes.

However, all the attention-based methods focus on one type of image regions for question-image association (i.e., either free-form image regions or detection boxes), and each of them have limitations on solving certain types of questions. In contrast, in order to better utilize the complementary information from both types of image regions, our proposed approach effectively integrates both attention mechanisms in a unified framework.

\begin{figure*}[ht!]
	\centering
	\includegraphics[width=0.88\linewidth]{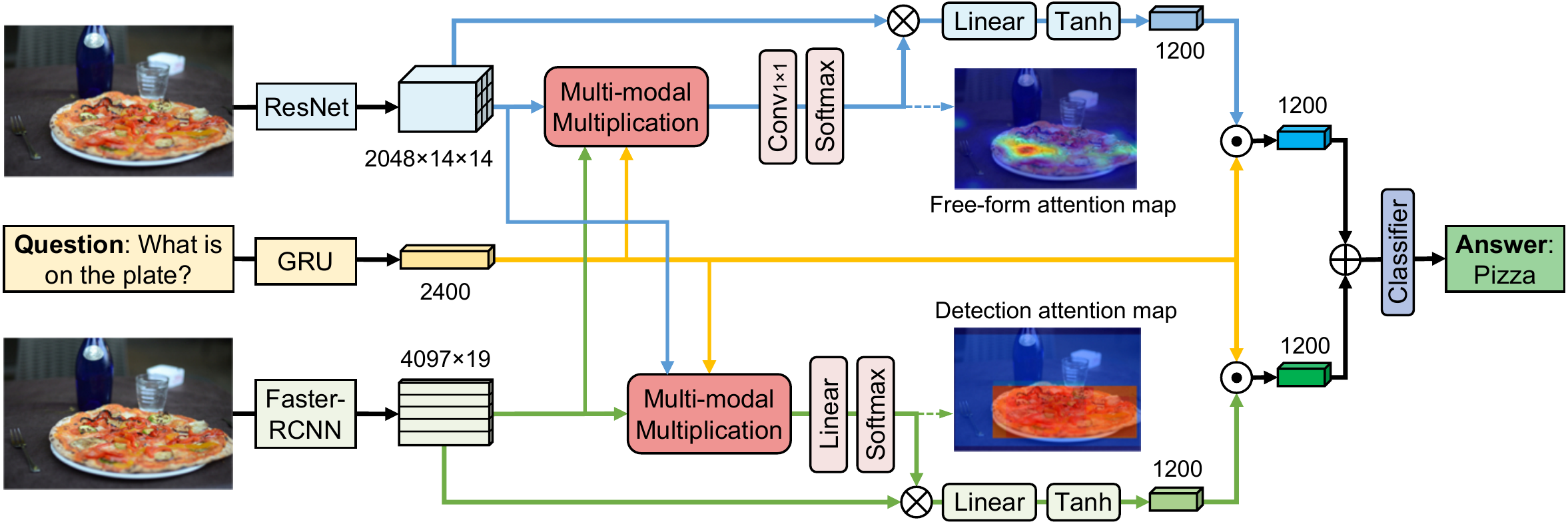}
	\caption{Illustration of the overall network structure for solving the VQA task. The network has two attention branches with the proposed multiplicative featucre embedding scheme, where one branch attends free-form image regions and another branch attends detection boxes for encoding question-related visual features.}
	\label{fig:bi-mlt}
\end{figure*}


\section{Method}

The overall structure of our proposed deep neural network is illustrated in Figure \ref{fig:bi-mlt}, which takes the question, the whole image, and the detection boxes as inputs, and learns to associate questions to free-form image regions and detection boxes simultaneously to infer the answer. 
Our proposed model for VQA consists of two co-attention branches for free-form image regions and for detection boxes, respectively. Before calculating the attention weights for each type of regions, the joint feature representations of the question, the whole-image visual features, and the detection box visual features are obtained via a multiplicative approach. In this way, the attention weights for each type of regions are learned from all the three inputs. Importantly, the joint multi-modal feature embedding has different parameters for each branch, which leads to better answering accuracy. The resulting visual features from the two branches are fused with the question representation and then added to obtain the final question-image embedding. A multi-class linear classifier is adopted for obtaining the final predicted answer.

The input feature encoding will be introduced in Section \ref{31}. Section \ref{32} introduces the branch with visual attention mechanism for associating free-form image regions to the input questions with our multiplicative feature embedding scheme, and Section \ref{33} describes the another branch with detection based attention mechanism. The final answer prediction will be introduced in Section \ref{34}.

\subsection{Input feature encoding}\label{31}


\textbf{Encoding whole-image features}. We utilize an ImageNet pre-trained ResNet-152 \cite{he2016deep} for learning image visual features. The input image to ResNet is resized to $448\times 448$ and the $2048\times14\times14 $ output feature map of the last convolution layer is used to encode the whole-image visual features $R \in \mathcal{R}^{2048\times14\times14}$, where $14\times14$ is its spatial size and $2048$ is the number of visual feature channels.

\textbf{Encoding detection-box features.} We adopt the Faster-RCNN \cite{ren2015faster} framework to obtain object detection boxes in the input image. For all the object proposals and their associated detection scores generated by Faster-RCNN, the non-maximum suppression with an intersection over union (IoU) 0.3 is applied and the top-ranked 19 detection boxes are chosen as the detection-box features for our overall framework. The 4096-dimensional visual features of Faster-RCNN's fc7 layer concatenated with boxes' detection scores are utilized to encode the visual feature of each detection box. Let $D = [f_1,\cdots, f_{19}]  \in \mathcal{R}^{4097\times19}$ denote the visual features of the top-ranked 19 detection boxes.                                                                                                                                                                                                                                                                                                                                                                                                                                                                                                                                                                                                                                                              

\textbf{Encoding question features.} The Gated Recurrent Unit (GRU) \cite{cho2014learning} is adopted to encode the question features, which show  its effectiveness in recent VQA methods \cite{lu2016hierarchical,kim2016hadamard}. A GRU cell consists of an update gate $z$ and a reset gate $r$. Given a question $q = [q_1,..., q_T]$, where $q_t$ is the one-hot vector of at position $t$, and $T$ is the length of the question. We convert each word $q_t$ into a feature vector via a linear transformation $x_t = W_eq_t$. At each time step, the word feature vector $q_t$ is sequentially fed into the GRU to encode the input question. At each step, the GRU updates the update gate $z_t$ and reset gate $r_t$ and outputs a hidden state $h_t$. The GRU update process operates as
\begin{align}
	&z_{t}=\sigma(W_{z}x_{t}+U_{z}h_{t-1}+b_{z}), \\
	&r_{t}=\sigma(W_{r}x_{t}+U_{r}h_{t-1}+b_{r}), \\
	&\tilde{h}_{t}=\tanh(W_{h}x_{t}+U_{h}(r_{t}\circ h_{t-1})+b_{h}), \\
	&h_{t}=z_{t}\circ h_{t-1}+(1-z_{t}) \circ \tilde{h}_{t},
\end{align}
where $\sigma$ represents the sigmoid activation function. The weight matrices $W$, $U$ and bias vector $b$ are learnable parameters of the GRU.

We take the final hidden state $h_T$ as the question embedding, i.e., $Q = h_T \in \mathcal{R}^{k} $, where $k$ denotes the embedding length. Following \cite{li2016visual,kim2016hadamard}, we utilize the pre-trained skip-thought model {\cite{kiros2015skip} to initialize the embedding matrix $W_e$ in our question language model. Since the skip-thought model is previously trained on large text corpus, we are able to transfer external language-based knowledge to our VQA task by fine-tuning the GRU from the initial point.

\subsection{Learning to attend free-form region visual features with multiplicative embedding}\label{32}

Our proposed framework has two branches, one for attending question-related free-form image regions to learn whole-image features and the other for attending detection-box regions to learn question-related detection features. For each attention branch, it takes question feature $Q$, whole-image feature $R$ and detection-box feature $D$ as inputs, and outputs the question-attended visual features $v_1$ and $v_2$ for question answering.

Our free-form region attention branch tries to associate the input question to relevant regions of the input image. There is no restriction to the attended regions, which are of free-form and are able to capture the global visual context and attributes of the image. Unlike existing VQA methods that fuse question and image features via simple concatenation \cite{shih2016look} or addition \cite{lu2016hierarchical} to guide the calculation of the attention weights, each of our attention branch fuses all three types of input features via a multiplicative feature embedding scheme for  utilizing full information from the inputs. The network structure for attending visual features with our multiplicative feature embedding scheme is show in Figure \ref{fig:mmf}(a).

Given the question embedding $Q \in \mathcal{R}^k$, the whole-image representation $R \in \mathcal{R}^{2048 \times 14 \times 14}$, and the detection representation $D \in \mathcal{R}^{4097 \times 19}$, we first embed them to a $1200$-dimensional common space via the following equations,
\begin{align}
R_1 & = \tanh(W_{r_1} R + b_{r_1}), \\
D_1 & = \frac{1}{19} \cdot {\bf 1} (\tanh(W_{d_1}D+b_{d_1})^T) \label{eq:d1}, \\
Q_1 & = \tanh(W_{q_1}Q+b_{q_1}),
\end{align}
where $W_{r_1} \in  \mathcal{R}^{1200 \times 2408}$, $W_{d_1} \in  \mathcal{R}^{1200 \times 4097}$, $W_{q_1} \in  \mathcal{R}^{1200 \times k}$ are the learnable weight parameters, $b_{r_1}$, $b_{d_1}$, $b_{q_1}$ $ \in  \mathcal{R}^{1200}$ are the bias parameters, ${\bf 1} \in \mathcal{R}^{19}$ represents an all-1 vector, and $\tanh$ is the hyperbolic tangent function. For learning the attention weights for the whole-image feature $R$, the transformed detection features are averaged across all detection boxes following Eq. (\ref{eq:d1}) before feature fusion. After mapping all input features into the $1200$-dimensional common space, the detection feature $D_1 \in \mathcal{R}^{1200}$ and question feature $Q_1 \in \mathcal{R}^{1200}$, are spatially replicated to a $14\times 14$ grid to form $\tilde{D}_1$ and $\tilde{R}_1$, which match the spatial size of the whole-image feature $R_1 \in \mathcal{R}^{1200\times 14 \times 14}$.

The joint feature representation $C_1$ of the three inputs is obtained by element-wise multiplication (Hadamard product) of $\tilde{Q}_1$, $R_1$ and $\tilde{D}_1$, and followed by a $L_2$ normalization to constrain the magnitude of the representation,
\begin{align}
C_1 & =  \mathrm{Norm_2} (\tilde{Q}_1 \circ R_1 \circ \tilde{D}_1), \label{eq:c1} 
\end{align}
where $\circ$ indicates element-wise multiplication. The free-form attention map is then obtained by convolving the joint feature representation $C_1$ with a $1\times 1$ convolution followed by a softmax operation over the $14\times 14$ grid,
\begin{align}
a_1 & = \mathrm{softmax}(W_{c_1} * C_1+b_{c_1}),
\end{align}
where $W_{c_1} \in \mathcal{R}^{1200 \times 1 \times 1}$ and $b_{c_1} \in  \mathcal{R}^{1200}$ are the learnable convolution kernel parameters. The attended whole-image feature over all spatial locations can be calculated by
\begin{align}
v_1  =  \sum_{i}^{14\times 14} a_1(i) R_1(i),
\end{align}
which can represent the attended whole-image visual feature that are most related to the input question.


\begin{figure}[th!]
	\centering
	\includegraphics[width=1.0\linewidth]{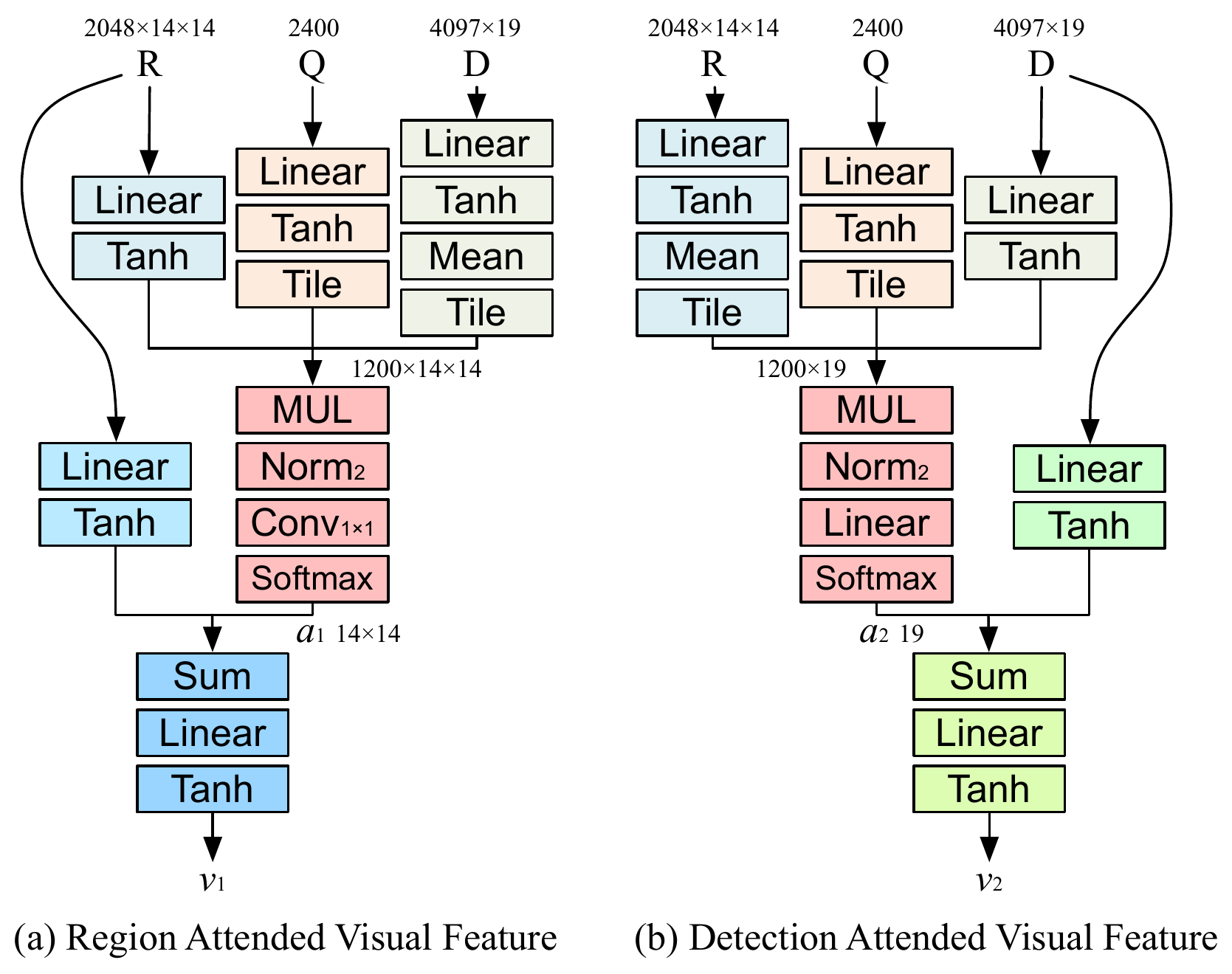}
	\caption{Learning to attend visual features with multi-modal multiplicative feature embedding for (a) free-form image regions and for (b) detection boxes.}
	\label{fig:mmf}
\end{figure}


\subsection{Learning to attend detection-box visual features with multiplicative embedding}\label{33}

Our second branch focuses on learning question-related detection features by attending the detection-box features with joint input feature representation. Similar to the first branch, we fuse the question representation, the whole-image features, and the detection-box features for learning the attention weights for detection boxes. Unlike previous detection based attention methods \cite{shih2016look,li2016visual}, our proposed attention mechanism integrates whole-image features for better understanding the overall image contents. The structure for learning the joint feature representation is shown in Figure \ref{fig:mmf}(b).

Similar to the joint representation for free-form image region attention, given the question embedding $Q \in \mathcal{R}^k$, the image representation $R \in \mathcal{R}^{2048 \times 14 \times 14}$, the detection representation $D \in \mathcal{R}^{4097 \times 19}$, we transform each representation to a common semantic space, and then obtain the $1200$-dimensional joint feature representation $C_2$ as 
\begin{align}
D_2 & = \tanh(W_{d_2}D+b_{d_2}), \\
R_2 & = \frac{1}{196} \cdot {\bf 1} (\tanh(W_{r_2}R+b_{r_2})^T), \\
Q_2 & = \tanh(W_{q_2}Q+b_{q_2}), \\
C_2 & =  \mathrm{Norm_2} (\tilde{Q}_2 \circ \tilde{R}_2 \circ D_2), \label{eq:c2}
\end{align}
where $W_{d_2}$, $W_{r_2}$, $W_{q_2}$ , $b_{d_1}$, $b_{r_1}$, $b_{q_2}$ are the learnable parameters for linear transformation. The transformed question feature $Q_2$ and whole-image feature $R_2$ are replicated across the number dimension to match the dimension of the transformed detection features $D_2 \in  \mathcal{R}^{1200\times 19}$ before calculating the joint representation $C_2$ following Eq. (\ref{eq:c2}).

The final attended detection representation $v_2$ over all detection boxes is defined as
\begin{align}
a_2 & = \mathrm{softmax}(W_{c_2} C_2+b_{c_2}), \\
v_2 & =  \sum_{i}^{19}a_2(i) {D_2}(i), 
\end{align}
where $W_{c_2} \in  \mathcal{R}^{19 \times 1200}$ and $b_{c_2} \in \mathcal{R}^{19}$ are the learnable parameters of learning the attention weights $a_2$ for the detection boxes.

\subsection{Learning for answer prediction}\label{34}

Similar to existing VQA approaches \cite{antol2015vqa,lu2016hierarchical,fukui2016multimodal}, we model VQA as a multi-class classification problem. Given the attended whole-image feature $v_1$, attended detection feature $v_2$ with the input question feature $Q$, question-image joint encodings are obtained by element-wise multiplication of the transformed question features and the attended features from both branch,
\begin{align}
	h_r & = v_1 \circ \tanh(W_{hr}Q+b_{hr}), \\
	h_d & = v_2 \circ \tanh(W_{hd}Q+b_{hd}),
\end{align}
where $W_{hr}$, $W_{hd}$, $b_{hr}$, $b_{hd}$ are the learnable parameters for transforming the input question feature $Q$. The reason why we choose different question transformation parameters for the two branches is that the attended visual features from the two different branches captures different information from the input image. One is from attended free-form region features and is able to capture the global context and attributes of the scene. The another is from attended detection features and is able to extract information about foreground objects.

After merging question-image encodings from the two branches via addition, a linear classifier is trained for final answer prediction,
\begin{align}
p_{ans} &= \mathrm{softmax}(W_p(h_r+h_d)+b_{p}), \label{eq:19}
\end{align}
where $W_p$ and $b_p$ are the classifier parameters, and $p_{ans}$ represents the probability of the final answer prediction.


\section{Experiments}

\subsection{Datasets and evaluation metrics}

We evaluate the proposed model on two public datasets, the VQA \cite{antol2015vqa} and COCO-QA \cite{ren2015exploring} datasets.

\textbf{VQA} dataset \cite{antol2015vqa} is based on Microsoft COCO image data \cite{lin2014microsoft}.
The dataset consists of 248,349 training questions, 121,512 validation questions and 244,302 testing questions, generated on a total of 123,287 images. There are three types of questions including \textit{yes/no}, \textit{number} and \textit{other}. For each question, 10 free-response answers are provided. We take the top 2000 most frequent answers as the candidate outputs for learning the classification model similar to \cite{kim2016multimodal}, which covers 90.45\% answers in the training and validation sets.

\textbf{COCO-QA} dataset \cite{ren2015exploring} is created based on the caption annotations of Microsoft COCO dataset \cite{lin2014microsoft}. There are 78,736 training samples and 38,948 testing samples, respectively. It contains four types of questions, \textit{object}, \textit{number}, \textit{color} and \textit{location}. All of the answers are single-words and considered as the valid answers, namely 430 answers, which are used for the possible answer classification.

Since we formulate VQA as a classification task, we use the accuracy metric to measure the performances of different models on both two datasets. In particular, for the VQA dataset, a predicted answer is regarded as correct if it matches more than three in ten ground truth answers. WUPS calculates the similarity between two words based on their common subsequence in the taxonomy tree. In addition, Wu-Palmer similarity (WUPS) \cite{wu1994verbs} is also reported for the COCO-QA dataset. Same as the previous work \cite{lu2016hierarchical,li2016visual}, we report the WUPS scores with the thresholds of 0.9 and 0.0.

\subsection{Implementation details}

For encoding questions, the length of questions is fixed to 26 and each word embedding is a vector of size 620. The hidden state of GRU is set to 2400. Given the question, image and detection representations, the joint feature embedding of these inputs $h$ is set as 1200. Following \cite{kim2016hadamard}, we take two glimpses of each attention map, that is to say, for each branch, another set of attention weights $a_1$ is trained for whole-image features $R_1$ in the first branch and another set of $a_2$ for detection features $D_2$ in the second branch. The two sets of attended features are concatenated in each branch as the final feature.

We implement our model with the Torch library. The RMSProp method is used for training our network with an initial learning rate of $3\times 10^{-4}$, a momentum of 0.99 and a weight-decay of $10^{-8}$. The batch size is set to 300, and is trained for 250,000 iterations. The validation process is performed every 10,000 iterations with early stopping when validation accuracy stops improving for more than 5 validations. Dropout is applied after every linear transformation and gradient clipping techniques are used for regularization in GRU training. Multi-GPU parallel technology is adopted to accelerate the training process.

\subsection{Comparison with state-of-the-arts}

\begin{table*}[th!] \label{table1}
	\centering 
	\small 

	\begin{tabular}{{l}*{10}{c}}
		
		\toprule
		\multirow{3}{*}{} &
		\multicolumn{5}{c}{Test-dev} & 
		\multicolumn{5}{c}{Test-std} 	\\
		
		\cmidrule(lr){2-6} 	\cmidrule{7-11} 	&
		\multicolumn{4}{c}{Open-Ended} &  	MC &
		\multicolumn{4}{c}{Open-Ended} &  	MC \\
		
		\cmidrule(lr){2-5} 	\cmidrule(lr){6-6} 	  \cmidrule(lr){7-10} 	\cmidrule(lr){11-11}
		\textbf{Method}	& All & Y/N &  Num.  & Other  & All  
		& All & Y/N & Num. & Other &  All \\
		
		\midrule
		LSTM Q+I \cite{antol2015vqa}  	 & 53.74 & 78.94 & 35.24 & 36.42 & 57.17
		& 54.06 & 79.01 & 35.55 & 36.80 & 57.57   \\
		iBOWING \cite{zhou2015simple}  	 & 55.72 & 76.55 & 35.03 & 42.62 & 61.68
		& 55.89 & 76.76 & 34.98 & 42.62 & 61.97   \\
		DPPnet \cite{noh2016image}			& 57.22 & 80.71 & 37.24 & 41.69 & 62.48
		& 57.36 & 80.28 & 36.92 & 42.24 & 62.69   \\
		FDA \cite{ilievski2016focused}  	& 59.24 & 81.14 & 36.16 & 45.77 & 64.01
		& 59.54 & 81.34 & 35.67 & 46.10 & 64.18   \\
		DMN+ \cite{xiong2016dynamic}  & 60.30 & 80.50 & 36.80 & 48.30 & -
		& 60.36 & 80.43 & 36.82 & 48.33 & -   \\
		\midrule
		Region Sel. \cite{shih2016look}   & - & - & - & - & 62.44
		& - & - & - & -  & 62.43   \\
		QRU \cite{li2016visual}						& 60.72 & 82.29 & 37.02 & 47.67 & 65.43
		& 60.76 & - & - & - & 65.43  \\
		\midrule
		SMem \cite{xu2016ask}		  	 		& 57.99 &  80.87 & 37.32 & 43.12 & -
		& 58.24 & 80.80 & 37.53 & 43.48 & -   \\
		SAN \cite{yang2016stacked}		  	 & 58.70 & 79.30 & 36.60 & 46.10 & -
		& 58.85 & 79.11 & 36.41 & 46.42 & -   \\
		MRN \cite{kim2016multimodal}	& 61.68 & 82.28 & 38.82 & 49.25 & 66.15
		& 61.84 & 82.39 & 38.23 & 49.41 & 66.33   \\
		HieCoAtt \cite{lu2016hierarchical} & 61.80 & 79.70 & 38.70 & 51.70 & 65.80
		& 62.06 & 79.95 & 38.22 & 51.95 & 66.07   \\
		VQA-Machine \cite{wang2017machine} & 63.10 & 81.50 & 38.40 & 53.00 & 67.70
		& 63.30 & 81.40 & 38.20 & 53.20 & 67.80   \\
		MCB \cite{fukui2016multimodal}	 & 64.70 & 82.50 & 37.60 & 55.60 & 69.10
		& - & - & - & - & -   \\
		MLB \cite{kim2016hadamard}			& 64.89 & \textbf{84.13} & 37.85 & 54.57 & -
		&  65.07 & \textbf{84.02} & 37.90 & 54.77 & 68.89   \\
		\midrule
		Dual-MLB (our baseline)											 & 65.12 & 83.32 & 39.96 & 55.26 & 69.55
		& - & - & - & - & -   \\
		\textbf{Dual-MFA} (ours)	& \textbf{66.01}& 83.59 & \textbf{40.18} & \textbf{56.84}  & \textbf{70.04}
		& \textbf{66.09} & 83.37 & \textbf{40.39} & \textbf{56.89} & \textbf{69.97}   \\		
		\bottomrule	
		
	\end{tabular}
	\caption{Evaluation results by our proposed method and compared methods on the VQA dataset.}
	\label{tab:vqa}
\end{table*}

Table \ref{tab:vqa} shows results on the VQA test set for both open-ended and multiple-choice tasks by our proposed approach and compared methods. The approaches shown in Table \ref{tab:vqa} are trained on the train+val split of the VQA dataset and evaluated on the test split, where the \textit{test-dev} set is normally used for validation and the \textit{test-std} set for standard testing. The models in the first part of Table \ref{tab:vqa} are based on simple joint question-image feature embedding methods. Models in the second part of Table \ref{tab:vqa} employ detection-based attention mechanism and models in the third part use free-form region-based attention mechanism. We only compare results of the single models since most approaches do not adopt the model ensemble strategy. We can see that our final model (denoted as Dual-MFA, where MFA stands for Multiplicative Feature Attention) improves the state-of-the-art MLB approach \cite{kim2016hadamard} from 65.07\% to 66.09\% for the open-ended task, and from 68.89\% to 69.97\% for the multiple-choice task on the test-std set. Specifically, in the question types of \textit{number} and \textit{other}, our proposed approach brings 2.49\% and 2.12\% improvements on the test-std set. QRU \cite{li2016visual} is the state-of-the-art detection-based attention method, and our model significantly outperforms it by 5.33\% on the test-std set. We also compare our approach with a baseline model Dual-MLB, which consists of two attention branches based on the MLB  tensor fusion module. The baseline model fuses the question and free-form image region embedding by MLB in one branch, and fuses question and image detection  embedding by MLB in another branch. The baseline Dual-MLB also outperforms MLB by 0.23\% on the test-dev set, which shows that our improvements are not only from the integration of the two attention mechanisms, but are also caused by effective joint feature embedding of question, image, and detection features.

Table \ref{tab:coco} compares our approach with the state-of-the-art approaches on the COCO-QA test set. Our final model Dual-MFA improves the state-of-the-art HieCoAtt \cite{lu2016hierarchical} from 65.40\% to 66.49\%. In particular, our model achieves an improvement of 2.99\% for the question type \textit{color}. Similar to the results on the VQA dataset, our model significantly outperforms the state-of-the-art detection-based attention method QRU by 3.49\%.

\subsection{Ablation study}
In this section, we conduct ablation experiments to study effectiveness of individual component designs in our model. Table \ref{tab:val} shows the results of baseline models replacing different components in our model, which are trained on the VQA training set and tested on the validation set. Following other compared approaches, the test set is not used in the study due to online submission restrictions. Specifically, we compare different multi-modal feature embedding designs and investigate the roles of the two attention mechanisms.

\begin{table}[h!]
	\centering 
	\small 
	
	\begin{tabular}{{l}*{5}{c}}
		\toprule
		
		\textbf{Method}  & \textbf{Validation}  \\
		\midrule
		MFA-MUL*	 		   & 59.01  \\	
		MFA-ADD	 				& 56.34  \\ 
		\midrule
		MFA-Norm*		& 59.18  \\	
		MFA	w/o Norm 						 & 59.01  \\  
		MFA-Power	 		   & 58.93  \\   
		\midrule
		Dual-MFA-ADD*	  & 59.82  \\
		Dual-MFA-MUL	   & 58.35  \\	
		Dual-MFA-CAT	 	& 58.53  \\		
		\midrule
		MFA-D*	 			    & 55.57   \\	
		QRU-D \cite{li2016visual}		   		& 53.99  \\ 
		MLB-D \cite{kim2016hadamard}	   				& 54.89  \\ 
		\midrule
		MFA-R*	 			  & 59.18  \\	
		MLB-R \cite{kim2016hadamard}   			  & 57.40  \\	
		MUTAN \cite{benyounescadene2017mutan}	 	 & 58.16  \\	
		\midrule
		Dual-MLB 	 		& 59.07  \\
		\textbf{Dual-MFA}* (full model) 	& \textbf{59.82}  \\	
		\bottomrule
		
	\end{tabular}
	\caption{Ablation study on the VQA dataset, where ``*'' denotes our model's design.}
	\label{tab:val}
\end{table}

\begin{table*}[t!]
	\centering 
	\small 
		
	\begin{tabular}{{l}*7{c}} 
		\toprule
		
		\textbf{Method}	 & All  & Obj. & Num. & Color & Loc.  & WUPS0.9 & WUPS0.0 \\
	
		\midrule
		2VIS+BLSTM \cite{ren2015exploring}	& 55.09 & 58.17 & 44.79 & 49.53 & 47.34   & 65.34 & 88.64 \\	
		IMG-CNN \cite{ma2016learning}			& 58.40 & - & - & - & -                                         & 68.50 & 89.67 \\	
		DDPnet \cite{noh2016image}				   &  61.16	& - & - & - & -                                         & 70.84  & 90.61 \\
		SAN \cite{yang2016stacked}				    &  61.60 & 65.40  & 48.60 & 57.90 & 54.00 & 71.60  & 90.90 \\
		QRU \cite{li2016visual}	 						   & 62.50 & 65.06 & 46.90 & 60.50 & 56.99   & 72.58 & 91.62 \\		 
		HieCoAtt \cite{lu2016hierarchical}		 & 65.40 &	68.00 & 51.00 & 62.90 & 58.80   & 75.10 & 92.00 \\	
		\midrule
		\textbf{Dual-MFA} (ours)     & \textbf{66.49} &\textbf{68.86} &\textbf{51.32} &\textbf{65.89} &\textbf{58.92}    &\textbf{76.15} & \textbf{92.29}  \\		 
		\bottomrule
		
	\end{tabular}
	\caption{Evaluation results by our proposed method and compared methods on the COCO QA dataset.}
	\label{tab:coco}
\end{table*}

\begin{figure*}[t!]
	\tiny
	\begin{tabular}{c@{\hspace{-2.5mm}}c}
		&
		\begin{tabular}{c@{\hspace{2mm}}c@{\hspace{2mm}}c@{\hspace{2mm}}c@{\hspace{2mm}}c@{\hspace{2mm}}c}
			
			\vspace{0.01in}
			\includegraphics[height=0.9in,width=1.3in]{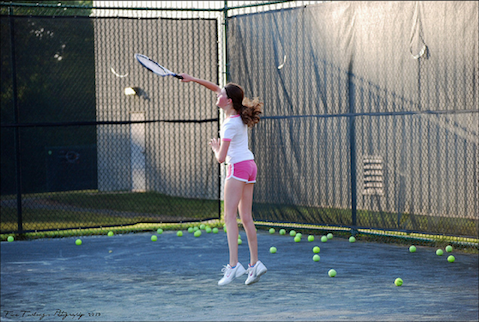}&
			\includegraphics[height=0.9in,width=1.3in]{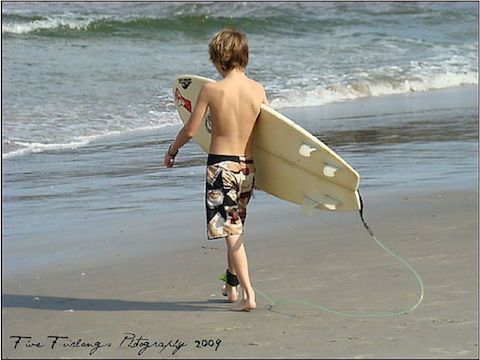}&
			\includegraphics[height=0.9in,width=1.3in]{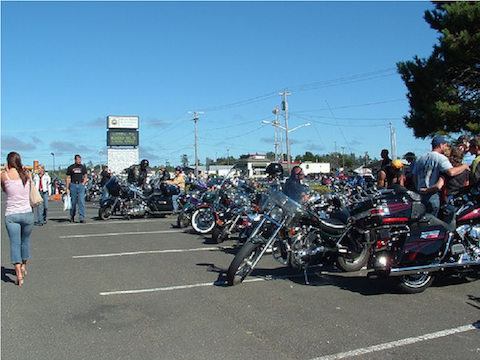}&
			\includegraphics[height=0.9in,width=1.3in]{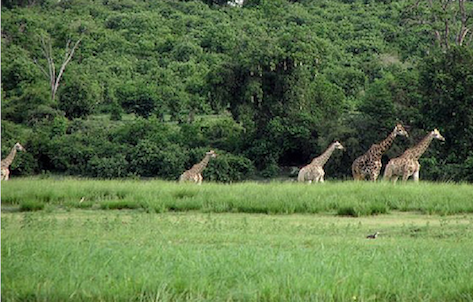}&	
			\includegraphics[height=0.9in,width=1.3in]{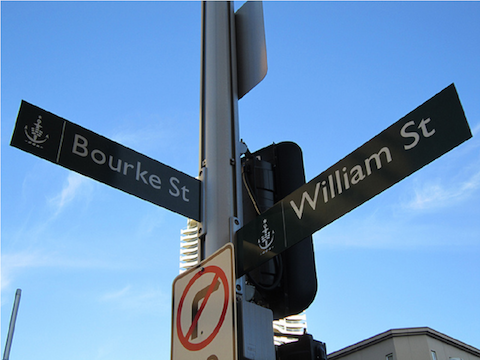}\\
			
			\textbf{Q}: {What sport is this?} & 
			\textbf{Q}: {What is the color of the surfboard?} & 
			\textbf{Q}: {Is it a sunny day?} & 
			\textbf{Q}: {How many giraffes are there?} &
			\textbf{Q}: {What does the red circle sign mean?} \\
			\vspace{0.01in}
			\textbf{A}: {\color{ForestGreen} tennis} & 
			\textbf{A}: {\color{ForestGreen} white} & 
			\textbf{A}: {\color{ForestGreen} yes} & 
			\textbf{A}: {\color{ForestGreen} 5} &
			\textbf{A}: {\color{red} no parking} \\ 
			
			\includegraphics[height=0.9in,width=1.3in]{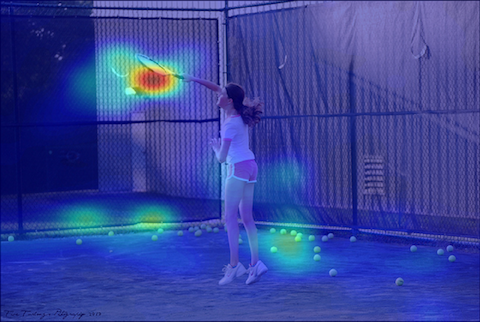}&
			\includegraphics[height=0.9in,width=1.3in]{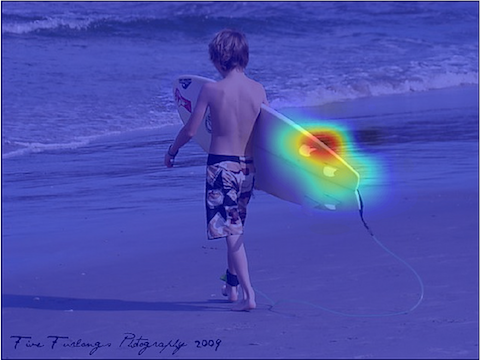}&
			\includegraphics[height=0.9in,width=1.3in]{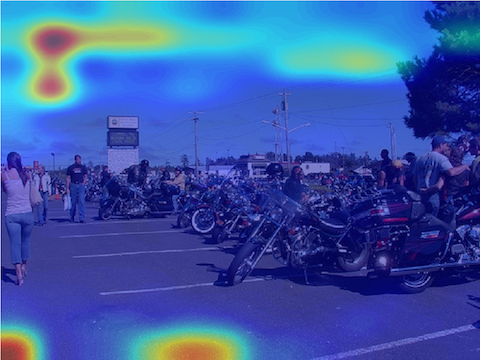}&
			\includegraphics[height=0.9in,width=1.3in]{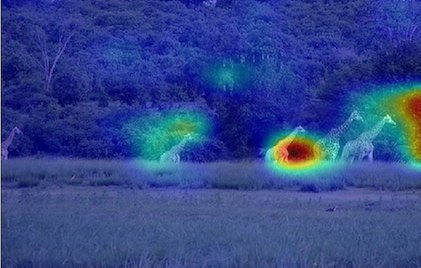}&
			\includegraphics[height=0.9in,width=1.3in]{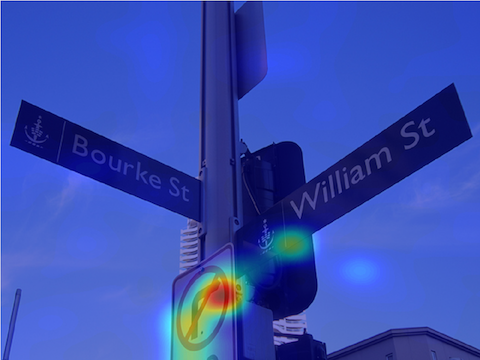}\\	
			
			\includegraphics[height=0.9in,width=1.3in]{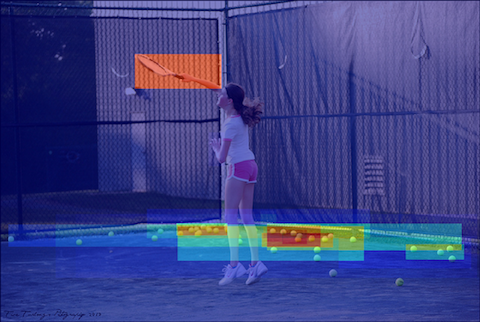}&
			\includegraphics[height=0.9in,width=1.3in]{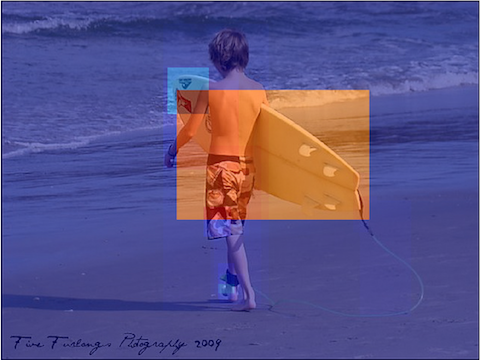}&
			\includegraphics[height=0.9in,width=1.3in]{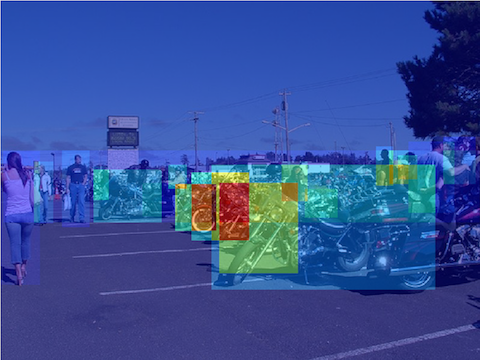}&
			\includegraphics[height=0.9in,width=1.3in]{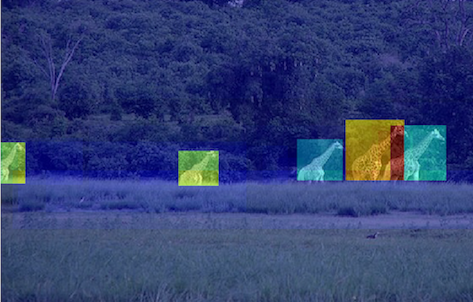}&
			\includegraphics[height=0.9in,width=1.3in]{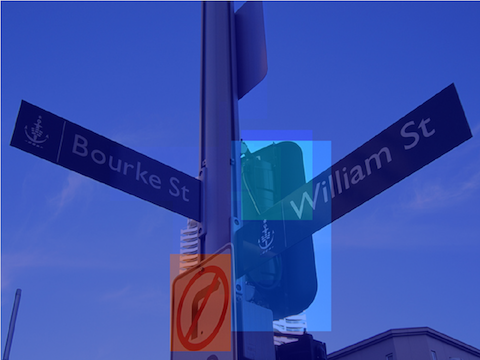}\\
			
			{\normalsize(a) } & {\normalsize(b) } & {\normalsize(c)} & {\normalsize(d) }  & {\normalsize(e) }
		\end{tabular}
	\end{tabular}
	\caption{Visualization examples on the VQA test-dev set. (First row) input images. (Second row) free-form region based attention maps. (Third row) detection-based attention maps.}
	\label{fig:vis}
\end{figure*}

The first part of Table \ref{tab:val} shows different element-wise operations used for joint embedding of three input features. Element-wise multiplication (denoted as MFA-MUL) in Eq. (\ref{eq:c1}) performs better than element-wise addition (denoted as MFA-ADD) by 3.67\%.· The second part of Table \ref{tab:val} shows $L_2$ normalization (denoted as MFA-Norm) in joint feature embedding works better than the model with unsigned power operation of $1/3$ (denoted as MFA-Power) and the model without normalization (denoted as MFA w/o Norm). For fusing outputs $h_r$ and $h_d$ from the two attention branches, element-wise addition in Eq. (\ref{eq:19}) achieves better performance than element-wise multiplication (denoted as Dual-MFA-MUL) and concatenation of two vectors (denoted as Dual-MFA-CAT).

The third and fourth parts of Table \ref{tab:val} compare our multi-modal multiplicative feature embedding for a single attention mechanism (MFA-R or MFA-D) with detection-based attention baseline models (QRU-D and MLB-D) and region-based attention baseline models (MUTAN and MLB-R), which replace our multiplicative feature embedding with QRU \cite{li2016visual} or MLB \cite{kim2016hadamard} respectively. The MUTAN \cite{benyounescadene2017mutan} approach is also used for comparison. Results show that our models with a single attention branch outperform all compared detection-based and region-based baselines. Finally, we compare our final model Dual-MFA with the baseline model Dual-MLB. Our Dual-MFA model benefits from its effective multi-modal multiplicative feature embedding scheme and achieves an improvement of 0.75\%. 

The parameter size of our final full model is 82.6M, while the best region-based model (MFA-R) is 61.7M, and the best detection-based model (MFA-D) is 56.8M. As the parameters in the language model and the classifier are shared by two attention branches, parameters of our full model increase in a minor degree.

\subsection{Qualitative evaluation}

We visualize some co-attention maps generated by our model in Figure \ref{fig:vis} and present five examples from the VQA test set. Figures \ref{fig:vis}(a) and (b) show that our model attends to the corresponding image regions with two attention branches, which leads to correct answers with higher confidence. There are also cases where only one attention branch is able to attend the correct image region to obtain the correct answer.
In Figure \ref{fig:vis}(c), the free-form region based attention map is able to attend the blue sky and dry ground, while the detection-based attention map fails as there are no pre-given detection boxes covering these image regions. In Figure \ref{fig:vis}(d), the detection-based attention map has higher weights on all five giraffes to generate the correct answer, while the free-form attention map attends incorrect regions. A failure case is also presented in Figure \ref{fig:vis}(e). The model fails to generate the correct answer as the classification label ``no turning right'' does not exit in the training set despite attending the correct image region.

\section{Conclusion}

In this paper, we propose a novel deep neural network with co-attention mechanism for visual question answering. The deep model contains two branches for visual attention which aim to select the free-form image regions and detection boxes most related to the input question. We generate visual attention weights by a novel multiplicative embedding scheme, which fuses the question, whole-image and detection-box features effectively. Ablation study demonstrates the effectiveness of individual components of our proposed model.
Experimental results on two large VQA datasets show that our proposed model outperforms state-of-the-art approaches.

\section*{Acknowledgement}

This work was supported in part by National Natural Science Foundation of China under Grant No. 61532010 and No. 61702190, National Basic Research Program of China (973 Program) under Grant No. 2014CB340505, and SHMEC (16CG24),
in part by SenseTime Group Limited, 
in part by the General Research Fund sponsored by the Research Grants Council of Hong Kong (Nos. CUHK14213616, CUHK14206114,
CUHK14205615, CUHK419412, CUHK14203015, CUHK14239816, CUHK14207814), in part by the Hong Kong Innovation and Technology Support Programme Grant ITS/121/15FX. 
We also thank Tong Xiao, Kun Wang and Kai Kang for helpful discussions.

\fontsize{9.5pt}{10.5pt} \selectfont
\bibliographystyle{aaai}
\bibliography{aaai18_ref}

\begin{thebibliography}{}

\bibitem[\protect\citeauthoryear{Antol \bgroup et al\mbox.\egroup
  }{2015}]{antol2015vqa}
Antol, S.; Agrawal, A.; Lu, J.; Mitchell, M.; Batra, D.; Lawrence~Zitnick, C.;
  and Parikh, D.
\newblock 2015.
\newblock Vqa: Visual question answering.
\newblock In {\em ICCV}.

\bibitem[\protect\citeauthoryear{Ben-Younes \bgroup et al\mbox.\egroup
  }{2017}]{benyounescadene2017mutan}
Ben-Younes, H.; Cad{\`{e}}ne, R.; Thome, N.; and Cord, M.
\newblock 2017.
\newblock Mutan: Multimodal tucker fusion for visual question answering.
\newblock In {\em ICCV}.

\bibitem[\protect\citeauthoryear{Cho \bgroup et al\mbox.\egroup
  }{2014}]{cho2014learning}
Cho, K.; van Merri{\"{e}}nboer, B.; G{\"{u}}l{\c c}ehre, {\c C}.; Bahdanau, D.;
  Bougares, F.; Schwenk, H.; and Bengio, Y.
\newblock 2014.
\newblock Learning phrase representations using rnn encoder--decoder for
  statistical machine translation.
\newblock In {\em EMNLP}.

\bibitem[\protect\citeauthoryear{Fukui \bgroup et al\mbox.\egroup
  }{2016}]{fukui2016multimodal}
Fukui, A.; Park, D.~H.; Yang, D.; Rohrbach, A.; Darrell, T.; and Rohrbach, M.
\newblock 2016.
\newblock Multimodal compact bilinear pooling for visual question answering and
  visual grounding.
\newblock In {\em EMNLP}.

\bibitem[\protect\citeauthoryear{He \bgroup et al\mbox.\egroup
  }{2016}]{he2016deep}
He, K.; Zhang, X.; Ren, S.; and Sun, J.
\newblock 2016.
\newblock Deep residual learning for image recognition.
\newblock In {\em CVPR}.

\bibitem[\protect\citeauthoryear{Ilievski, Yan, and
  Feng}{2016}]{ilievski2016focused}
Ilievski, I.; Yan, S.; and Feng, J.
\newblock 2016.
\newblock A focused dynamic attention model for visual question answering.
\newblock {\em arXiv preprint arXiv:1604.01485}.

\bibitem[\protect\citeauthoryear{Karpathy and Fei-Fei}{2015}]{karpathy2015deep}
Karpathy, A., and Fei-Fei, L.
\newblock 2015.
\newblock Deep visual-semantic alignments for generating image descriptions.
\newblock In {\em CVPR}.

\bibitem[\protect\citeauthoryear{Kim \bgroup et al\mbox.\egroup
  }{2016}]{kim2016multimodal}
Kim, J.-H.; Lee, S.-W.; Kwak, D.; Heo, M.-O.; Kim, J.; Ha, J.-W.; and Zhang,
  B.-T.
\newblock 2016.
\newblock Multimodal residual learning for visual qa.
\newblock In {\em NIPS}.

\bibitem[\protect\citeauthoryear{Kim \bgroup et al\mbox.\egroup
  }{2017}]{kim2016hadamard}
Kim, J.-H.; On, K.~W.; Lim, W.; Kim, J.; Ha, J.-W.; and Zhang, B.-T.
\newblock 2017.
\newblock Hadamard product for low-rank bilinear pooling.
\newblock In {\em ICLR}.

\bibitem[\protect\citeauthoryear{Kiros \bgroup et al\mbox.\egroup
  }{2015}]{kiros2015skip}
Kiros, R.; Zhu, Y.; Salakhutdinov, R.~R.; Zemel, R.; Urtasun, R.; Torralba, A.;
  and Fidler, S.
\newblock 2015.
\newblock Skip-thought vectors.
\newblock In {\em NIPS}.

\bibitem[\protect\citeauthoryear{Li and Jia}{2016}]{li2016visual}
Li, R., and Jia, J.
\newblock 2016.
\newblock Visual question answering with question representation update (qru).
\newblock In {\em NIPS}.

\bibitem[\protect\citeauthoryear{Li \bgroup et al\mbox.\egroup
  }{2017a}]{li2017person}
Li, S.; Xiao, T.; Li, H.; Zhou, B.; Yue, D.; and Wang, X.
\newblock 2017a.
\newblock Person search with natural language description.
\newblock In {\em CVPR}.

\bibitem[\protect\citeauthoryear{Li \bgroup et al\mbox.\egroup
  }{2017b}]{li2017visual}
Li, Y.; Duan, N.; Zhou, B.; Chu, X.; Ouyang, W.; and Wang, X.
\newblock 2017b.
\newblock Visual question generation as dual task of visual question answering.
\newblock {\em arXiv preprint arXiv:1709.07192}.

\bibitem[\protect\citeauthoryear{Lin \bgroup et al\mbox.\egroup
  }{2014}]{lin2014microsoft}
Lin, T.-Y.; Maire, M.; Belongie, S.; Hays, J.; Perona, P.; Ramanan, D.;
  Doll{\'a}r, P.; and Zitnick, C.~L.
\newblock 2014.
\newblock Microsoft coco: Common objects in context.
\newblock In {\em ECCV}.

\bibitem[\protect\citeauthoryear{Lu \bgroup et al\mbox.\egroup
  }{2016}]{lu2016hierarchical}
Lu, J.; Yang, J.; Batra, D.; and Parikh, D.
\newblock 2016.
\newblock Hierarchical question-image co-attention for visual question
  answering.
\newblock In {\em NIPS}.

\bibitem[\protect\citeauthoryear{Ma, Lu, and Li}{2016}]{ma2016learning}
Ma, L.; Lu, Z.; and Li, H.
\newblock 2016.
\newblock Learning to answer questions from image using convolutional neural
  network.
\newblock In {\em AAAI}.

\bibitem[\protect\citeauthoryear{Malinowski, Rohrbach, and
  Fritz}{2015}]{malinowski2015ask}
Malinowski, M.; Rohrbach, M.; and Fritz, M.
\newblock 2015.
\newblock Ask your neurons: A neural-based approach to answering questions
  about images.
\newblock In {\em ICCV}.

\bibitem[\protect\citeauthoryear{Mostafazadeh \bgroup et al\mbox.\egroup
  }{2016}]{mostafazadeh2016generating}
Mostafazadeh, N.; Misra, I.; Devlin, J.; Mitchell, M.; He, X.; and Vanderwende,
  L.
\newblock 2016.
\newblock Generating natural questions about an image.
\newblock In {\em ACL}.

\bibitem[\protect\citeauthoryear{Noh, Hongsuck~Seo, and
  Han}{2016}]{noh2016image}
Noh, H.; Hongsuck~Seo, P.; and Han, B.
\newblock 2016.
\newblock Image question answering using convolutional neural network with
  dynamic parameter prediction.
\newblock In {\em CVPR}.

\bibitem[\protect\citeauthoryear{Ren \bgroup et al\mbox.\egroup
  }{2015}]{ren2015faster}
Ren, S.; He, K.; Girshick, R.; and Sun, J.
\newblock 2015.
\newblock Faster r-cnn: Towards real-time object detection with region proposal
  networks.
\newblock In {\em NIPS}.

\bibitem[\protect\citeauthoryear{Ren, Kiros, and
  Zemel}{2015}]{ren2015exploring}
Ren, M.; Kiros, R.; and Zemel, R.
\newblock 2015.
\newblock Exploring models and data for image question answering.
\newblock In {\em NIPS}.

\bibitem[\protect\citeauthoryear{Shih, Singh, and Hoiem}{2016}]{shih2016look}
Shih, K.~J.; Singh, S.; and Hoiem, D.
\newblock 2016.
\newblock Where to look: Focus regions for visual question answering.
\newblock In {\em CVPR}.

\bibitem[\protect\citeauthoryear{Wang \bgroup et al\mbox.\egroup
  }{2017}]{wang2017machine}
Wang, P.; Wu, Q.; Shen, C.; and Hengel, A. v.~d.
\newblock 2017.
\newblock The vqa-machine: Learning how to use existing vision algorithms to
  answer new questions.
\newblock In {\em CVPR}.

\bibitem[\protect\citeauthoryear{Wu and Palmer}{1994}]{wu1994verbs}
Wu, Z., and Palmer, M.
\newblock 1994.
\newblock Verbs semantics and lexical selection.
\newblock In {\em ACL}.

\bibitem[\protect\citeauthoryear{Xie, Shen, and Zhu}{2016}]{xie2016online}
Xie, L.; Shen, J.; and Zhu, L.
\newblock 2016.
\newblock Online cross-modal hashing for web image retrieval.
\newblock In {\em AAAI}.

\bibitem[\protect\citeauthoryear{Xiong, Merity, and
  Socher}{2016}]{xiong2016dynamic}
Xiong, C.; Merity, S.; and Socher, R.
\newblock 2016.
\newblock Dynamic memory networks for visual and textual question answering.
\newblock In {\em ICML}.

\bibitem[\protect\citeauthoryear{Xu and Saenko}{2016}]{xu2016ask}
Xu, H., and Saenko, K.
\newblock 2016.
\newblock Ask, attend and answer: Exploring question-guided spatial attention
  for visual question answering.
\newblock In {\em ECCV}.

\bibitem[\protect\citeauthoryear{Yang \bgroup et al\mbox.\egroup
  }{2016}]{yang2016stacked}
Yang, Z.; He, X.; Gao, J.; Deng, L.; and Smola, A.
\newblock 2016.
\newblock Stacked attention networks for image question answering.
\newblock In {\em CVPR}.

\bibitem[\protect\citeauthoryear{Ye \bgroup et al\mbox.\egroup
  }{2017}]{ye2017video}
Ye, Y.; Zhao, Z.; Li, Y.; Chen, L.; Xiao, J.; and Zhuang, Y.
\newblock 2017.
\newblock Video question answering via attribute-augmented attention network
  learning.
\newblock In {\em ACM Multimedia}.

\bibitem[\protect\citeauthoryear{Zhou \bgroup et al\mbox.\egroup
  }{2015}]{zhou2015simple}
Zhou, B.; Tian, Y.; Sukhbaatar, S.; Szlam, A.; and Fergus, R.
\newblock 2015.
\newblock Simple baseline for visual question answering.
\newblock {\em arXiv preprint arXiv:1512.02167}.

\bibitem[\protect\citeauthoryear{Zitnick and
  Doll{\'a}r}{2014}]{zitnick2014edge}
Zitnick, C.~L., and Doll{\'a}r, P.
\newblock 2014.
\newblock Edge boxes: Locating object proposals from edges.
\newblock In {\em ECCV}.

\end{thebibliography}

\end{document}